%% file: conference_101719.tex
\documentclass[conference]{IEEEtran}
\IEEEoverridecommandlockouts
\ifCLASSOPTIONcompsoc
    \usepackage[caption=false, font=normalsize, labelfont=sf, textfont=sf]{subfig}
\else
\usepackage[caption=false, font=footnotesize]{subfig}
\fi
\usepackage{cite}
\usepackage{amsmath,amssymb,amsfonts}
\usepackage{algorithmic}
\usepackage{graphicx}
\usepackage{textcomp}
\usepackage{xcolor}
\usepackage{float}
\def\BibTeX{{\rm B\kern-.05em{\sc i\kern-.025em b}\kern-.08em
    T\kern-.1667em\lower.7ex\hbox{E}\kern-.125emX}}
\begin{document}

\title{ECAvg: An Edge-Cloud Collaborative Learning Approach using Averaged Weights\\
}


\author{\IEEEauthorblockN{Atah Nuh Mih\IEEEauthorrefmark{1}, Hung Cao\IEEEauthorrefmark{1}, Asfia Kawnine\IEEEauthorrefmark{1}, Monica Wachowicz\IEEEauthorrefmark{1}\IEEEauthorrefmark{3}}

\IEEEauthorblockA{\IEEEauthorrefmark{1} \textit{Analytics Everywhere Lab, University of New Brunswick, Canada} \\
\IEEEauthorrefmark{3} \textit{RMIT University, Australia} \\\\}
}
\maketitle

\begin{abstract}
The use of edge devices together with cloud provides a collaborative relationship between both classes of devices where one complements the shortcomings of the other. Resource-constraint edge devices can benefit from the abundant computing power provided by servers by offloading computationally intensive tasks to the server. Meanwhile, edge devices can leverage their close proximity to the data source to perform less computationally intensive tasks on the data. In this paper, we propose a collaborative edge-cloud paradigm called ECAvg in which edge devices pre-train local models on their respective datasets and transfer the models to the server for fine-tuning. The server averages the pre-trained weights into a global model, which is fine-tuned on the combined data from the various edge devices. The local (edge) models are then updated with the weights of the global (server) model. We implement a CIFAR-10 classification task using MobileNetV2, a CIFAR-100 classification task using ResNet50, and an MNIST classification using a neural network with a single hidden layer. We observed performance improvement in the CIFAR-10 and CIFAR-100 classification tasks using our approach, where performance improved on the server model with averaged weights and the edge models had a better performance after model update. On the MNIST classification, averaging weights resulted in a drop in performance on both the server and edge models due to negative transfer learning. From the experiment results, we conclude that our approach is successful when implemented on deep neural networks such as MobileNetV2 and ResNet50 instead of simple neural networks.   
\end{abstract}


\input{Contents/1_intro}
\input{Contents/2_method}
\input{Contents/3_experiment-1}

\input{Contents/4_experiment-2}
\input{Contents/5_experiment-3}

\input{Contents/6_conclusion}

\section*{Acknowledgment}
This work is supported by the NBIF Talent Recruitment Fund (TRF2003-001). The equipment used in our experiments is supported by CFI Innovation Fund - Project ``Smart Campus Integration and Testing Lab"


\bibliographystyle{ieeetr}
\bibliography{reference.bib}

\end{document}

%% file: Contents/1_intro.tex
\section{Introduction}
In recent years, hardware developments have brought significant computational capabilities to edge devices. This potential has been leveraged by numerous vendors to provide hardware-accelerated devices for developing Edge AI solutions such NVIDIA's Jetson devices and Google's Coral Dev Board. These devices benefit from optimised hardware and software to improve inference time on edge devices. 

Albeit these hardware developments, edge devices are still constrained by memory, power consumption, and storage. These limitations motivate the use of cloud servers together with edge devices in a collaborative edge-cloud paradigm \cite{yao2022edge}. Several approaches have been explored, such as dedicating the edge device for data collection and inference, while storage and machine learning (ML) model development is assigned to the cloud. These methods use resource-efficient architectures such as MobileNetV2 \cite{sandler2018mobilenetv2}, EfficientNet \cite{tan2019efficientnet}, and YOLOv4 \cite{bochkovskiy2020yolov4}. The trained models are then deployed on the edge devices for inference in various applications, including pedestrian detection \cite{liu2022research}, autonomous vehicles \cite{chen2023edge}, and anomaly detection \cite{bibi2021edge}.

A common challenge faced by this approach is the viewpoint problem \cite{kukreja2019training}. Over time, the distribution of data received on the edge deviates from the data on which the inference model was trained, thereby reducing the model's performance. A new distribution, therefore, requires repeating the data collection, model training, and deployment processes.

On-device training can be leveraged to alleviate this viewpoint problem. The devices benefit from the availability of data collected onsite to build a model from the new data distribution. However, the resource constraints on the edge devices limit the amount of data they can use for training the models, as training on larger datasets is computationally intensive. These devices can, therefore, collaborate with cloud servers and benefit from their massive computational and storage capabilities \cite{cao2023fostering}. 

Several authors have proposed various edge-cloud collaboration approaches. Yang et al. \cite{yang2020big} proposed a cloud manufacturing system that uses gateways to process data on the edge and provide support for latency-sensitive applications, while the cloud performs computationally intensive data mining services. Hu et al. \cite{hu2020coedge} proposed CoEdge that greedily selects edge nodes (based on communication and computation resources) for deep learning and allocates deep neural networks over the edge-cloud environment while minimizing latency.



Federated Learning is another example of edge-cloud collaboration. Several authors \cite{wang2019adaptive} \cite{khan2020federated} \cite{su2021secure}  have proposed various federated learning mechanisms with edge devices as client nodes. In these works, the central server averages the weights from client models and updates the client devices with the global model. The server solely aggregates weights, has no data, and is therefore not directly involved in training the model. Our \textit{Collaborative Learning} approach differs from Federated Learning in that the server both aggregates the weights from client devices and trains on a central dataset. We average the pre-trained weights of the edge models into a global model, and leverage the computing power of the server to fine-tune the model on the combined dataset collected from the edge devices. The weights of this global model are then used to update the local edge models to improve their performance. 

In this paper, we propose a collaborative edge-cloud approach called ECAvg in which both the edge devices and the cloud servers participate in training. The edge devices benefit from their close proximity to the data source to pre-train local models on their respective data. Meanwhile, the server leverages its computing capability and transfer learning using pre-trained weights to fine-tune a global model on data aggregated from the edge devices \cite{cao2019analytics}. 
This collaborative paradigm is applicable in many consumer electronics contexts such as smart manufacturing, smart cities, or IoT. 

We aim to show the benefits of using averaged pre-trained weights from client edge devices to train a global model on aggregated data. We then study the training performance of the local edge models on their respective tasks when updated this fine-tuned global model.  

The main contributions of this work are as follows:
\begin{enumerate}
    \item We propose ECAvg - an edge-cloud collaborative learning approach that uses averaged weights to improve training performance on server and edge models.
    \item We implement our approach using three different ML model architectures and experiment with the proposed methods on CIFAR-10, CIFAR-100, and MNIST; and evaluate the performance of the models.
    \item We discuss the role of transfer learning in implementing our proposed ECAvg approach.
\end{enumerate}

Our paper is structured in the following order: Section I introduces a background the collaborative edge-cloud paradigm and discusses the relevant literature. Section II provides a mathematical formulation of our proposed method and Section III provides the implementation details. In Section IV, we experiment with three different datasets and model architectures and present their results. Section V provides an in-depth discussion of the success and shortcomings of our approach. Finally, we conclude our paper in Section VI.


%% file: Contents/2_method.tex
\section{Problem Formulation} \label{sec:prob_formulaton}
Consider an edge client-server setting with a central server and $M$ resource-constraint client devices. The server contains a dataset $\hat{D}$, and each client edge device $i \in M$ contains a dataset $D_i$ where $D_i \in \hat{D}$. 
$D_i$ consists of $n_i$ labeled samples and $\hat{D}$ consists of $N$ labeled samples such that
    \[N = \sum_{i=1}^M n_i\]

The client devices compute a model $h_\theta$, with parameters $\theta$ which maps X input space to Y label space. \textit{i.e.}
    \[h_\theta: X \rightarrow Y\]

The loss of the model on a sample $(x, y) \in (X, Y)$ is defined by
    \[l(h_\theta(x), y\]

For client $i$, the model learns a predictive function $f_i$ by calculating the average loss from $n_i$ samples in $D_i$.
    \begin{equation} \label{eqn:predictive_function}
        f_i (\theta) = \frac{1}{n_i} \sum_1^{n_i} l(h_\theta(x_i), y_i))
    \end{equation}

The final parameters $\theta$ computed by the client devices are sent to the edge server. The server finds an optimal model that minimizes the average of client losses. \textit{i.e.}
    \begin{equation} \label{eqn:server_model}
        h_{avg} = min_{\theta} \frac{1}{N} \sum_{i=1}^M f_i (\theta) = \frac{1}{N} \sum_{i=1}^M \sum l(h_\theta(x_i), y_i)
    \end{equation}
where $h_{avg}$ is optimal model with parameters $\theta_{avg}$

The server's computational resources offer an advantage over the resource-constraint edge devices. $h_{avg}$ can therefore be fine-tuned on the server's larger dataset $\hat{D}$ for better performance. The fine-tuned model parameters $\theta^*_{avg}$ are computed as follows:
    \begin{equation} \label{eqn:fine-tune}
        \theta^*_{avg} = \theta_{avg} - \mu g(\theta_{avg})
    \end{equation}
where $\mu$ is the learning rate and $g(\theta_{avg})$ is the stochastic gradient of the predictive function calculated on $\hat{D}$.

The server updates the client models with $\theta^*_{avg}$.

\section{Implementation}\label{sec:implementation}
Our setup consists of two A203 Mini PC edge devices and an Intel(R) Core(TM) i7-4790 CPU desktop, which we consider as a server.    

Let's assume $M_1$ trains on $D_1$ on a hypothetical edge device $E_1$, and $M_2$ trains on $D_2$ on a hypothetical edge device $E_2$. In practice, $E_1$ and $E_2$ are the same A203 Mini PC, but we make this distinction for ease of understanding. We summarize the process using the following steps:

\begin{enumerate}
    \item We train identical classifiers ($M_1$ and $M_2$) on the respective datasets ($D_1$ and $D_2$) for each client edge device.  
    \item After training, we transfer $M_1$ and $M_2$ to the server and average their weights. We build an identical global model, $\hat{M}$ and replace its weights with the averaged weights of the edge models. (Note that the architecture of the global model is adapted to accommodate the number of classes on the global dataset after the new weights have been added).
    \item By benefiting from prior knowledge learned in the edge models (transfer learning) and the server's computing resources, we fine-tune the global model on the full, $\hat{D}$. 
    \item The fine-tuned weights of the global model are then updated on the client models $M_1$ and $M_2$, and the models are once again trained on their respective datasets.
\end{enumerate}

\begin{figure}[!h]
     \centering
     \includegraphics[width=0.793\linewidth]{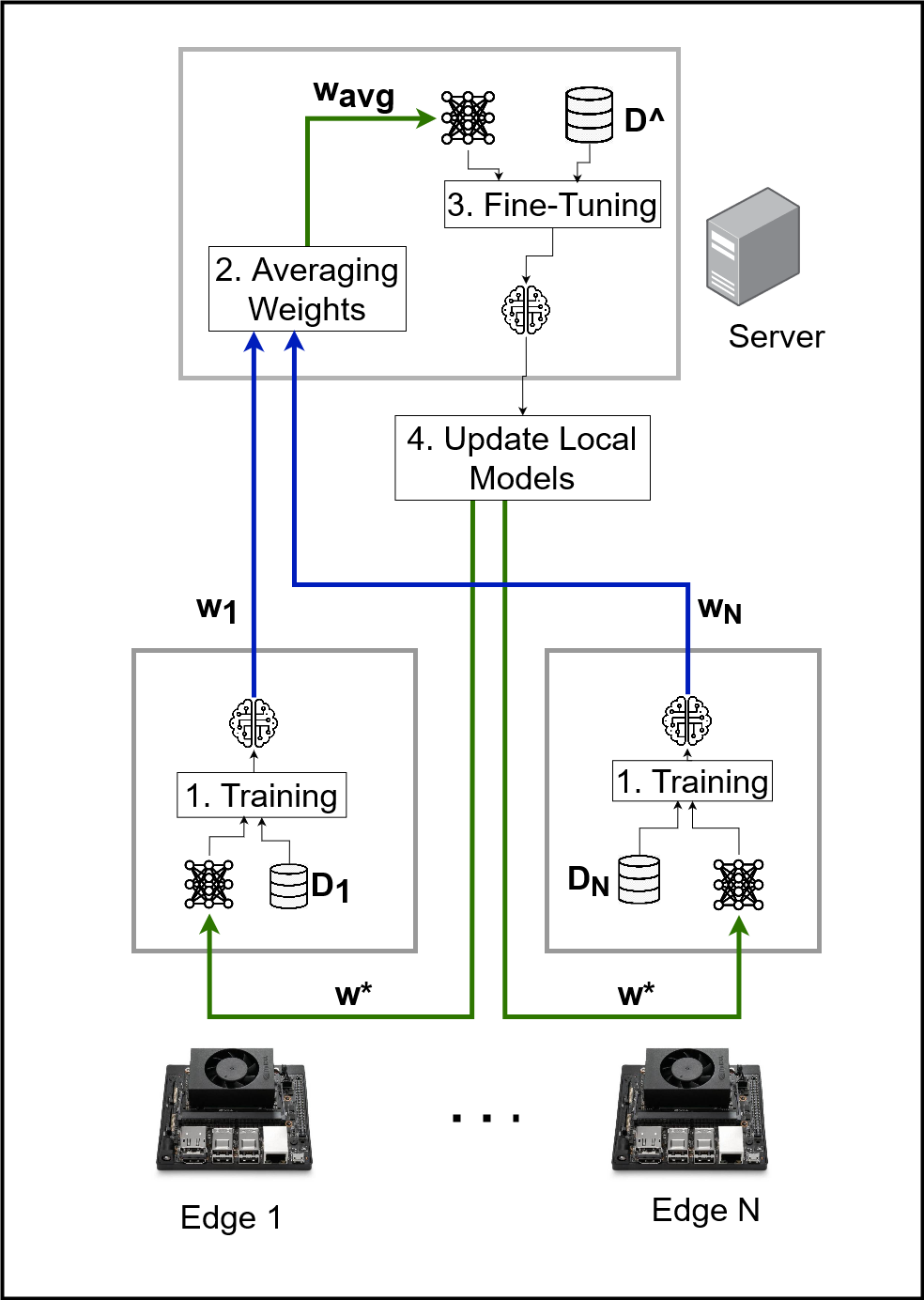}
     \caption{Overview of Proposed Method}
     \label{fig:overview}
 \end{figure}

Figure \ref{fig:overview} shows an overview of our proposed method.

%% file: Contents/3_experiment-1.tex
\section{Experiment}
We conduct three image classification experiments using the CIFAR-10 dataset \cite{cifar10}, CIFAR-100 dataset \cite{cifar10}, and the MNIST dataset \cite{lecun-mnisthandwrittendigit-2010}. We use MobileNetV2 and ResNet50 on the CIFAR-10 and CIFAR-100 datasets respectively to evaluate the performance of our Collaborative Learning approach with deep learning architectures. We then use a neural network with a single hidden layer on the MNIST dataset to evaluate the performance of our approach with a simple architecture.

\subsection{Experiment I - Applying ECAvg approach on MobileNetV2 with a small class size dataset (CIFAR10)}\label{sec:cifar10}
The CIFAR-10 dataset $\hat{D}$ consists of 10 classes of images: airplane, automobile, bird, cat, deer, dog, frog, horse, ship, and truck. We split the dataset into two smaller datasets of 5 classes each. Dataset 1 ($D_1$) contains airplane, automobile, bird, cat, and deer; and Dataset 2 ($D_2$) contains dog, frog, horse, ship, and truck.   

We build two identical MobileNetV2 classifiers $M_1$ and $M_2$, whose weights have been pre-trained on ImageNet. As both $M_1$ and $M_2$ are pre-trained on ImageNet, they have the same initialisation and represent $h_\theta$ in conjunction with the problem formulation in Section \ref{sec:prob_formulaton}. We perform the classification task as described in Section \ref{sec:implementation}, where we train the models on their respective datasets and transfer their weights to the server for averaging and fine-tuning. The fine-tuned server weights are then updated on the edge devices and the models are re-trained on their local datasets.

\subsubsection{Results and Observations}
\begin{figure*}[!h] 
    \centering
    \subfloat[Training on Edge Device $E_1$\label{fig_cifar10Model1}]{%
    \includegraphics[width=0.21\linewidth]{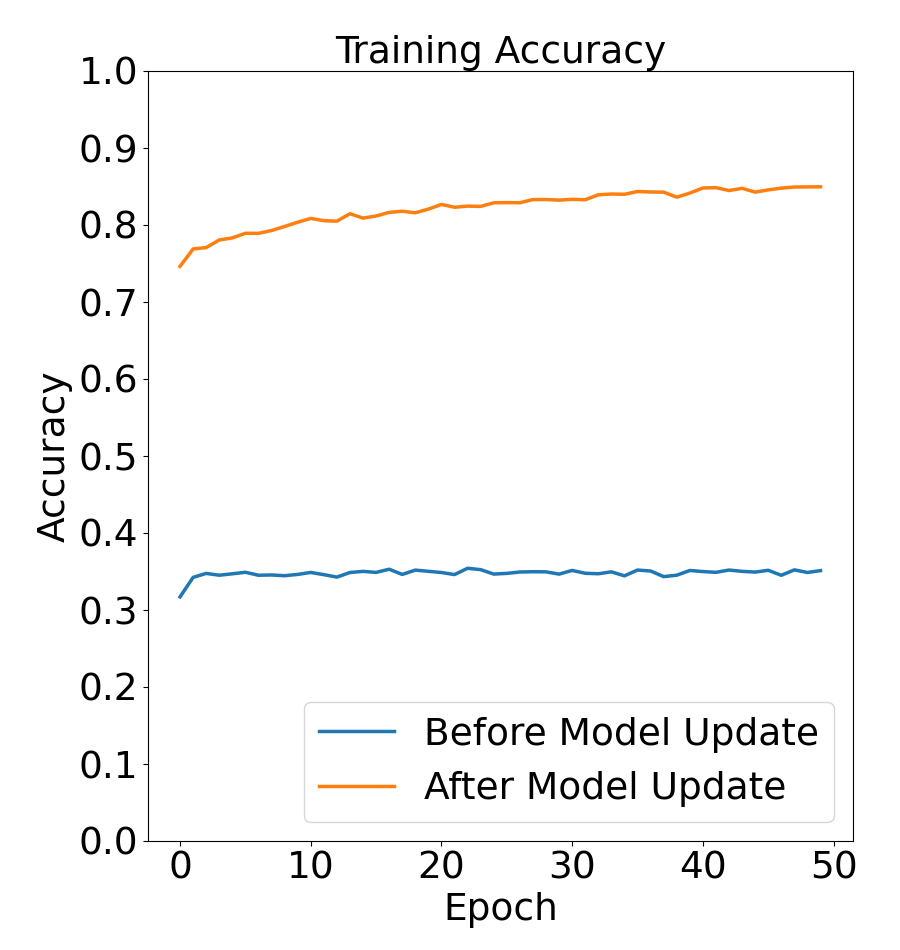}}\hspace{7mm}
    \subfloat[Training on Edge Device $E_2$\label{fig_cifar10Model2}]{%
        \includegraphics[width=0.21\linewidth]{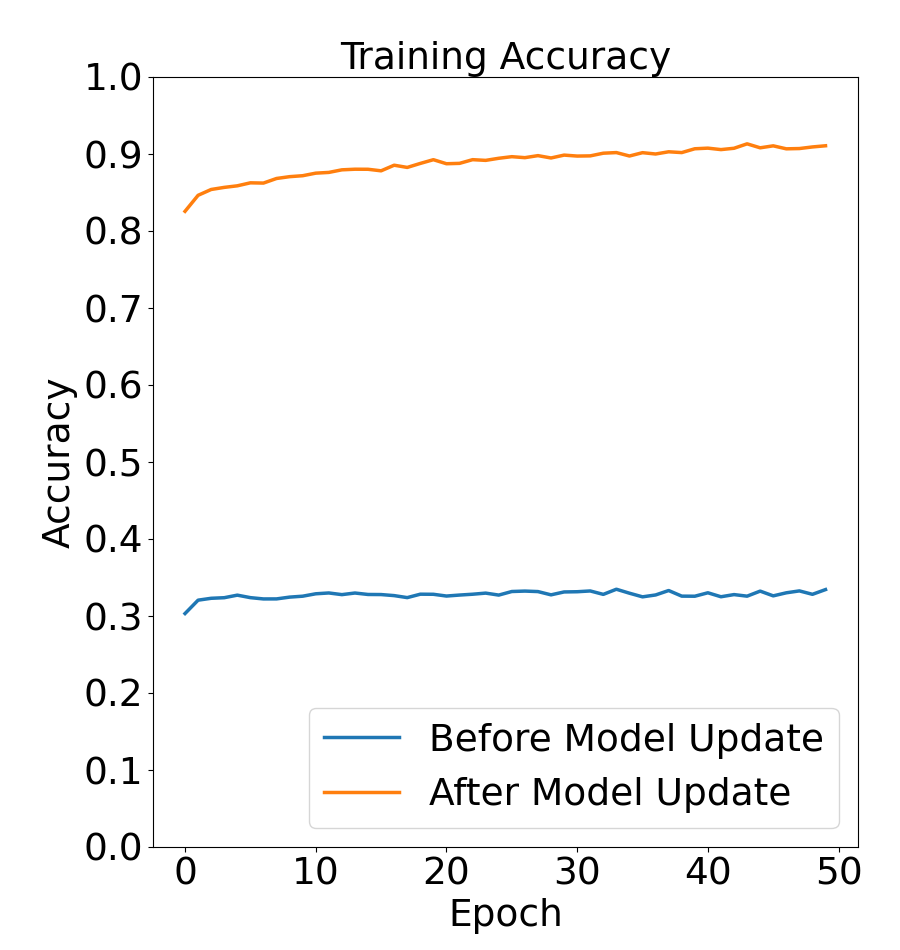}}\hspace{7mm}   
    \subfloat[Training on Server \label{fig_cifar10Server}]{%
         \includegraphics[width=0.21\linewidth]{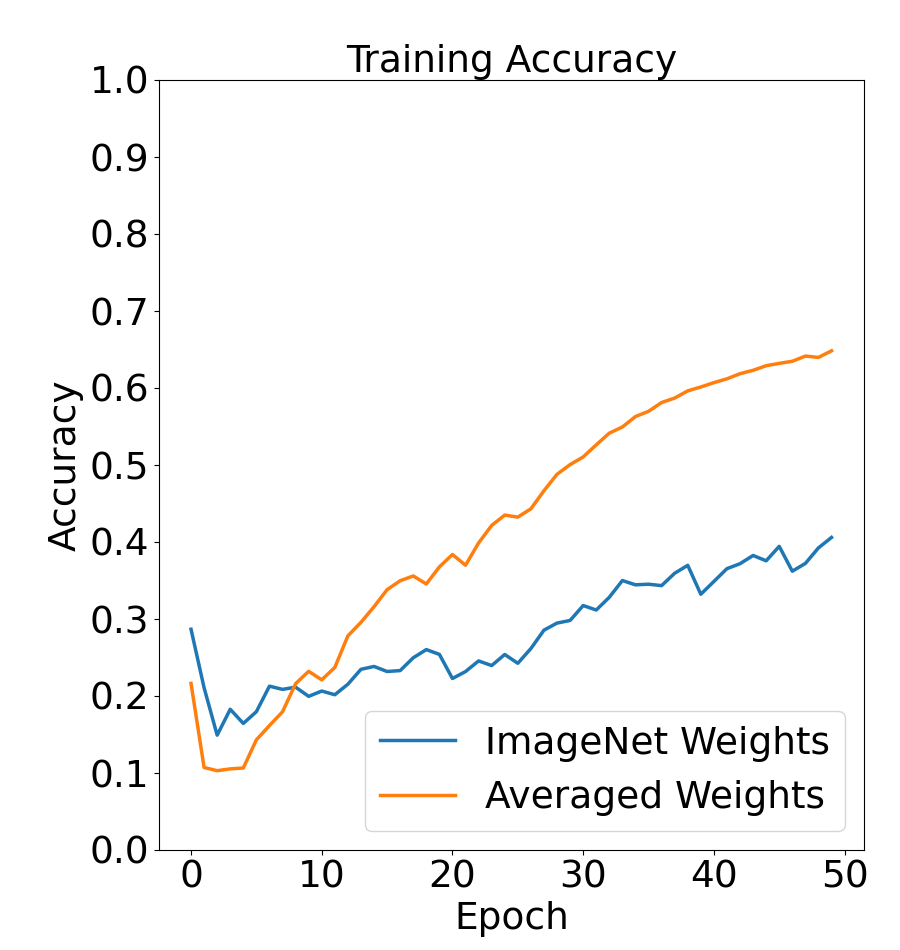}}
  \caption{Training performance on CIFAR-10 Dataset}
        \label{fig:cifar10_train_results}
\end{figure*}

Figure \ref{fig_cifar10Model1} and \ref{fig_cifar10Model2} show the training performance before and after the model update for edge devices $E_1$ and $E_2$. We notice significant improvements in training performance after the model update in both edge devices, with the models attaining high accuracy from the start and improving throughout training.    

We also evaluate the test performance of the edge models and server model in terms of accuracy, precision, recall, and F1 score and summarize the results in Table \ref{tab:cifar10_test}.

\begin{table}[h]
    \centering
    \caption{CIFAR-10 Test Results}
    \input{Contents/tables/cifar10_testing}   
    \label{tab:cifar10_test}
\end{table}

From Table \ref{tab:cifar10_test}, we observe a significant performance improvement on the edge devices after the model update. $M_1$ (on Edge 1) has the following test performance before the model update: 38.86\% accuracy, 40.96\% precision, 38.86\% recall, and 35.97\% F1 score. After the update, the following test performance is obtained: 81.16\% accuracy, 81.34\% precision, 81.16\% recall, and 80.90\% F1 score. 

Similarly, $M_2$ (on Edge 2) has the following test performance before the model update: 36.88\% accuracy, 38.09\% precision, 36.88\% recall, and 36.41\% F1 score. This performance improves to 87.40\% accuracy, 88.62\% precision, 87.40\% recall, and 87.11\% F1 score after the model update.

Fig \ref{fig_cifar10Server} shows the comparison of fine-tuning the server model with the averaged weights vs an identical model with weights pre-trained on ImageNet. Initially, the model with ImageNet weights has a better training accuracy, which does not improve significantly. The model with averaged weights has a low performance at the beginning of training, but this steadily improves throughout training and surpasses the model with ImageNet weights.



%% file: Contents/tables/cifar10_testing.tex
\resizebox{\linewidth}{!}{\begin{tabular}{llllll}
\hline \hline
\textbf{Device}   & \textbf{Setup}      & \textbf{Acc} & \textbf{Precision} & \textbf{Recall} & \textbf{F1 Score} \\
\hline
Edge 1          & Before update         & 0.3886            & 0.4096             & 0.3886          & 0.3597            \\
Edge 1          & After update            & \textbf{0.8116}   & \textbf{0.8134}    & \textbf{0.8116} & \textbf{0.8090}              \\
\hline
Edge 2          & Before update         & 0.3688            & 0.3809             & 0.3688          & 0.3641            \\
Edge 2          & After update            & \textbf{0.8740}   & \textbf{0.8862}    & \textbf{0.8740} & \textbf{0.8711}              \\
\hline
Server          & ImageNet weights           & 0.3660            & 0.4096             & 0.3660          & 0.3203            \\
Server          & Averaged weights              & \textbf{0.6696}   & \textbf{0.6946}    &\textbf{0.6696}  & \textbf{0.6641}   \\
\hline
\end{tabular}}

%% file: Contents/4_experiment-2.tex
\subsection{Experiment II - Applying ECAvg approach on ResNet50 with a large class size dataset (CIFAR100)} \label{sec:cifar100}
We carry out a similar experiment as in Section \ref{sec:cifar10} using the CIFAR-100 dataset \cite{cifar10}, consisting of 100 classes of images. In this experiment, we split the dataset into two distinct sub-datasets $D_1$ and $D_2$ containing 50 classes each.

We build two identical ResNet50 classifiers $M_1$ and $M_2$ for the two edge devices, where $M_1$ is trained on $D_1$ and $M_2$ is trained on $D_2$. Initially, the weights of the models are pre-trained on ImageNet before training on the edge devices. The trained models weights are transferred to the server and averaged. We build an identical server model $\hat{M}$ but adjust its architecture to accommodate training on the 100 classes of the complete CIFAR-100 dataset, $\hat{D}$. The weights of the server model are replaced with the averaged weights and the model is fine-tuned on $\hat{D}$. After fine-tuning, the edge models are updated with the fine-tuned weights of the server model. 

\subsubsection{Results and Observations}
\begin{figure*}[!h] 
    \centering
    \subfloat[Training on Edge Device $E_1$\label{fig_cifar100Model1}]{%
    \includegraphics[width=0.21\linewidth]{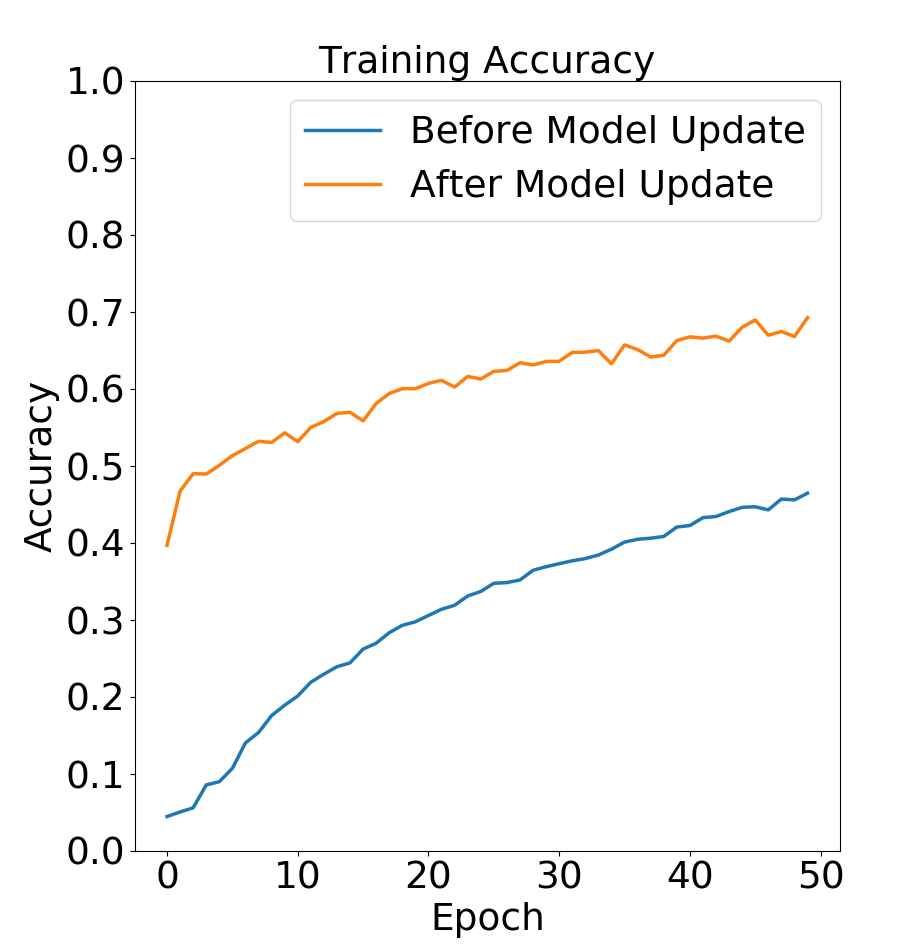}}\hspace{7mm}
    \subfloat[Training on Edge Device $E_2$\label{fig_cifar100Model2}]{%
        \includegraphics[width=0.21\linewidth]{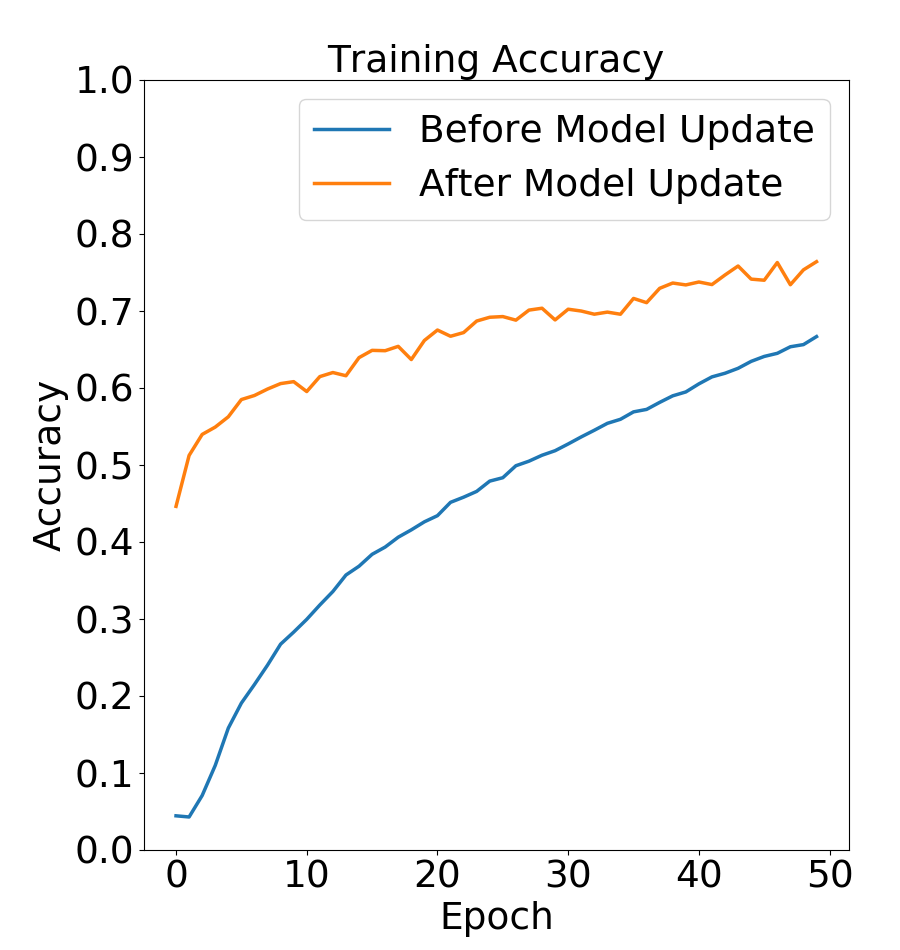}}\hspace{7mm}    
    \subfloat[Training on Server \label{fig_cifar100Server}]{%
         \includegraphics[width=0.21\linewidth]{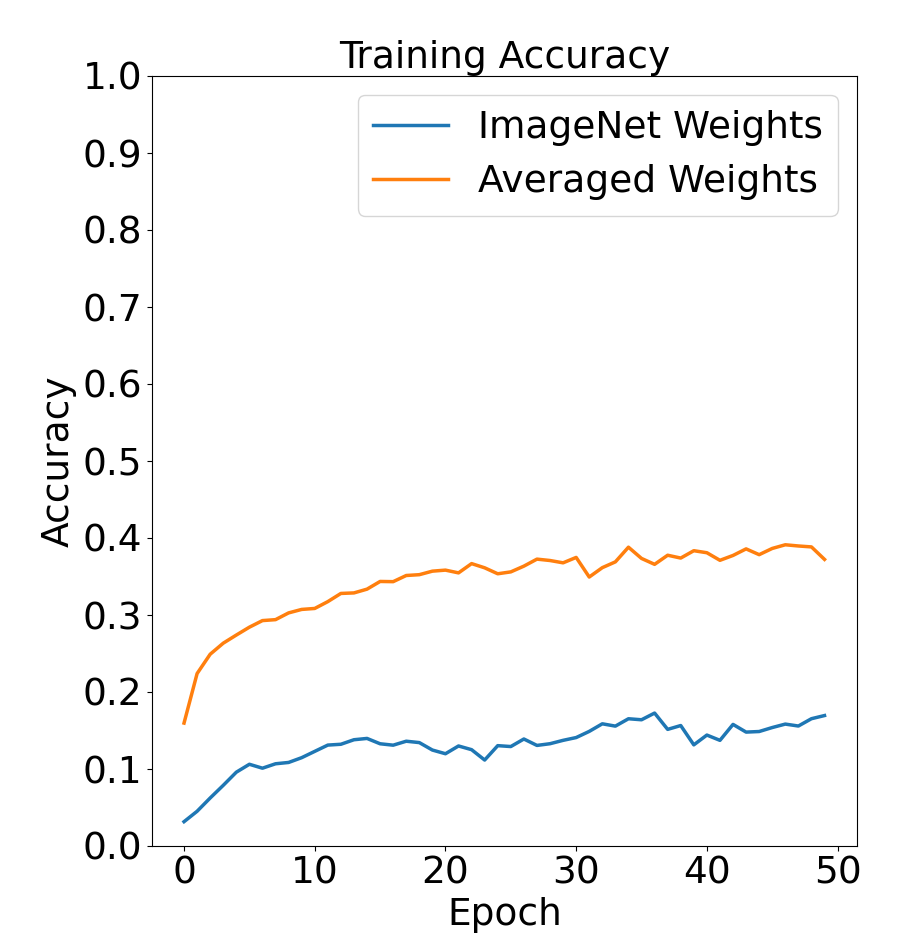}}
  \caption{Training performance on CIFAR-100 Dataset}
        \label{fig:cifar100_train_results}
\end{figure*}

We present the training results for the edge and server models in Fig \ref{fig:cifar100_train_results}. From Fig \ref{fig_cifar100Model1} and \ref{fig_cifar100Model2}, we observe that the training accuracy of the edge models improve after the model update, as compared to before the model update. The updated models also have a higher accuracy at the start of training, indicating that the edge models benefit from prior knowledge provided by the fine-tuned weights.

In Fig \ref{fig_cifar100Server}, we compare the performance of the server model with averaged weights against that of an identical model with weights pre-trained on ImageNet. The model with averaged weights has a better training accuracy throughout than the model pre-trained on ImageNet.

We further evaluate the performance of the models on their respective test datasets and present the results in Table \ref{tab:cifar100_test}. 

\begin{table}[h]
    \centering
    \caption{CIFAR-100 Test Results}
    \input{Contents/tables/cifar100_testing}   
    \label{tab:cifar100_test}
\end{table}

Prior to the model update on the edge devices, $M_1$ has a 39.64\% test accuracy, a 44.41\% precision, a 39.64\% recall, and a 39.27\% F1 score. $M_2$ has a 48.78\% test accuracy, a 49.55\% precision, a 48.78\% recall, and a 48.09\% F1 score.

After the model update, the performance of $M_1$ improves with a 51.56\% test accuracy, a 53.55\% precision, a 51.56\% recall, and a 51.27\% F1 score. Similarly, $M_2$ has an improved performance with a 51.80\% test accuracy, a 53.08\% precision, a 51.80\% recall, and a 51.40\% F1 score.

On the server, $\hat{M}$ with averaged weights has a 37.22\% test accuracy, a 39.38\% precision, a 37.45\% recall, and a 36.39\% F1 score. Meanwhile, $\hat{M}$ with ImageNet weights has a 21.00\% test accuracy, a 20.76\% precision, a 21.00\% recall, and a 19.10\% F1 score. These results indicate the advantage of using averaged weights over using weight pre-trained on another dataset.

%% file: Contents/tables/cifar100_testing.tex
\resizebox{\linewidth}{!}{\begin{tabular}{llllll}
\hline \hline
\textbf{Device}   & \textbf{Setup}      & \textbf{Acc} & \textbf{Precision} & \textbf{Recall} & \textbf{F1 Score} \\
\hline
Edge 1          & Before update         & 0.3964            & 0.4441             & 0.3964          & 0.3927            \\
Edge 1          & After update            & \textbf{0.5156}   & \textbf{0.5355}    & \textbf{0.5156} & \textbf{0.5127}              \\
\hline
Edge 2          & Before update         & 0.4878            & 0.4955             & 0.4878          & 0.4809            \\
Edge 2          & After update            & \textbf{0.5180}   & \textbf{0.5308}    & \textbf{0.5180} & \textbf{0.5140}              \\
\hline
Server          & ImageNet weights           & 0.2100            & 0.2076             & 0.2100          & 0.1910            \\
Server          & Averaged weights              & \textbf{0.3745}   & \textbf{0.3938}    &\textbf{0.3745}  & \textbf{0.3639}   \\
\hline
\end{tabular}}

%% file: Contents/5_experiment-3.tex
\subsection{Experiment III - Applying ECAvg approach on a simple neural netowrk with a small class size dataset (MNIST)}
The MNIST dataset $\hat{D}$ consists of 10 classes of images representing handrawn digits 0-9. Similarly, we divide the dataset into 2 subsets containing 5 classes each, where Dataset 1 ($D_1$) contains the digits 0-4, and Dataset 2 ($D_2$) contains the digits 5-9. We maintain the 5:5 splitting ratio for $D_1$ and $D_2$ in this experiment, but we consider experimenting with different splitting ratios in later experiments.

For the models $M_1$ and $M_2$, we build an identical neural network consisting of a single hidden layer and an output layer for the classification task.

\subsubsection{Results and Observations}
\begin{figure*}[!h] 
    \centering
    \subfloat[Training on Edge Device $E_1$\label{fig_mnistModel1}]{%
    \includegraphics[width=0.21\linewidth]{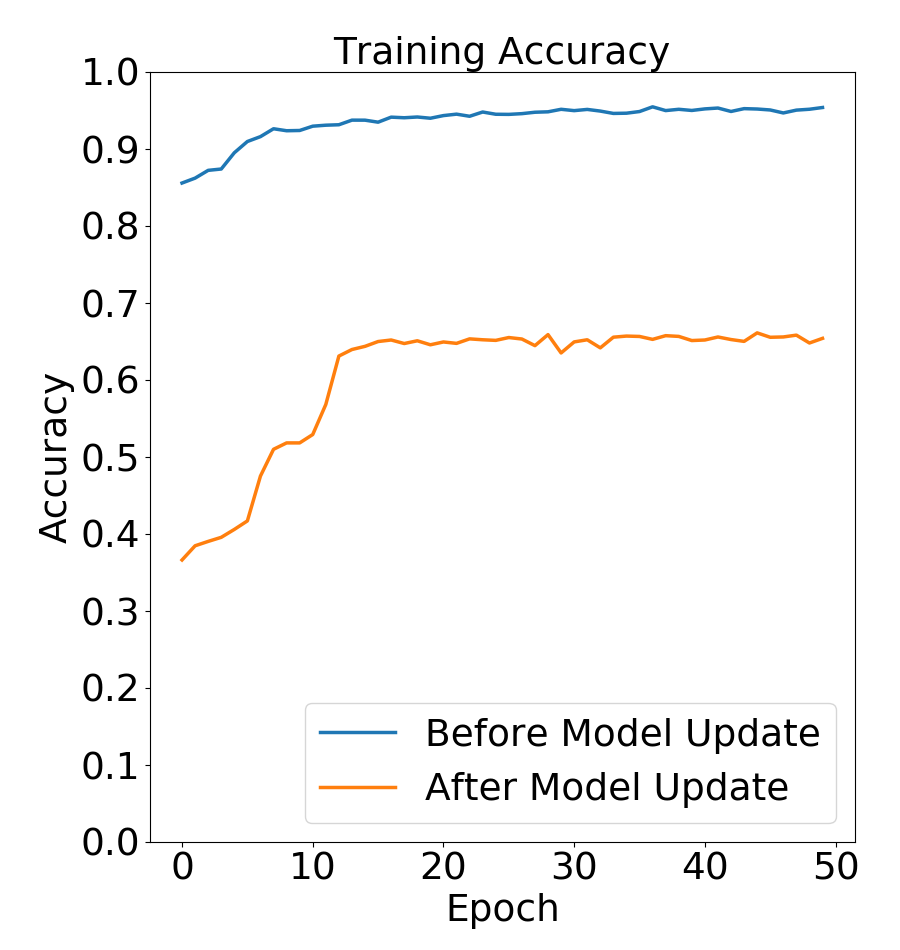}}\hspace{7mm}
    \subfloat[Training on Edge Device $E_2$\label{fig_mnistModel2}]{%
        \includegraphics[width=0.21\linewidth]{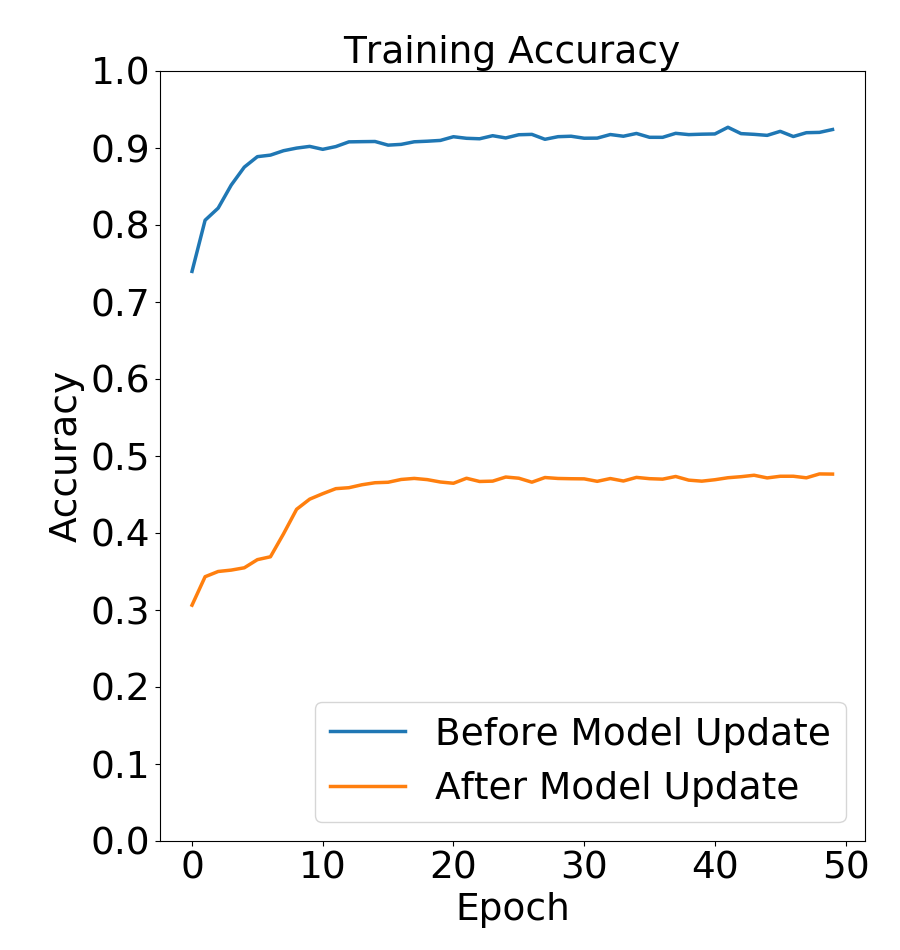}}\hspace{7mm}    
    \subfloat[Training on Server \label{fig_mnistServer}]{%
         \includegraphics[width=0.2\linewidth]{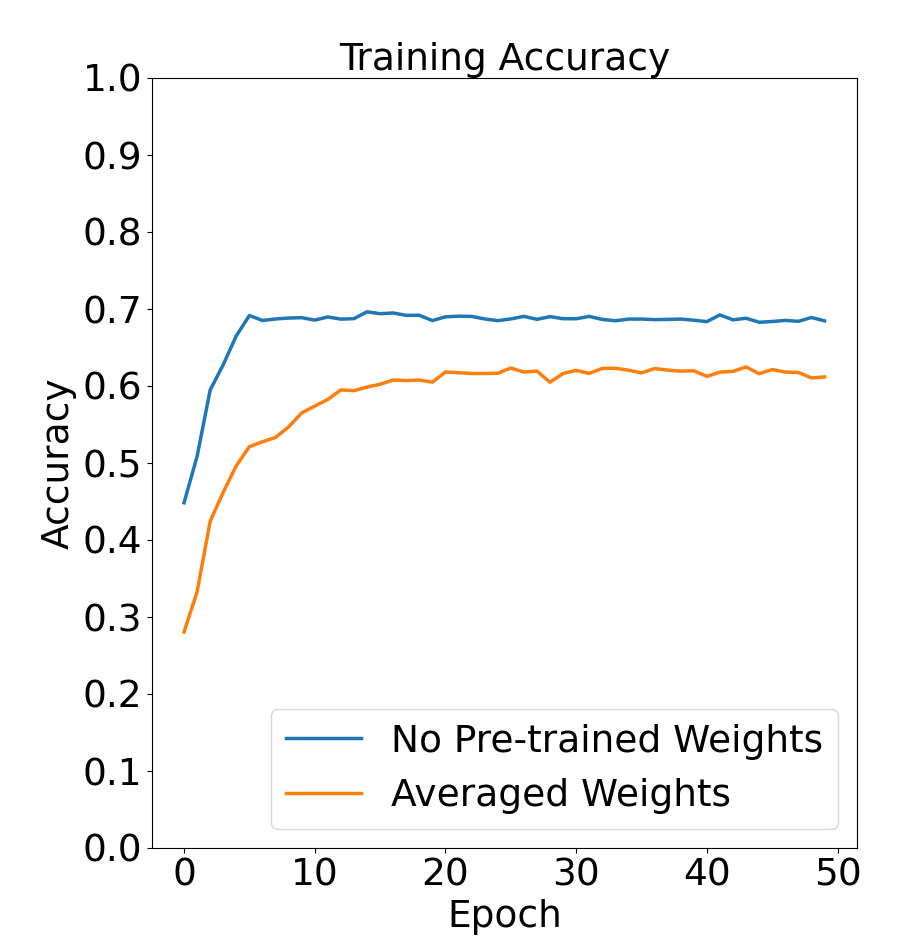}}
  \caption{Training performance on MNIST Dataset}
        \label{fig:mnist_train_results}
\end{figure*}

Figure \ref{fig_mnistModel1} and \ref{fig_mnistModel2} show the training results on both edge devices. We observe that the models on both edge devices perform better prior to the model update, and the update negatively affects the performance of the models. The same trend is observed while evaluating the models on the test datasets as shown in Table \ref{tab:mnist_test}. 

\begin{table}[h]
    \centering
    \caption{MNIST Test Results}
    \input{Contents/tables/mnist_testing}   
    \label{tab:mnist_test}
\end{table}

For $M_1$, the test accuracy drops from 97.22\% to 75.44\%, precision drops from 97.37\% to 67.84\%, recall drops from 97.22\% to 75.44\%, and F1 score drops from 97.25\% to 69.80\%. Similarly for $M_2$, the test accuracy drops from 92.40\% to 47.65\%, precision drops from 95.94\% to 48.36\%, recall drops from 95.68\% to 54.66\%, and F1 score drops from 95.71\% to 46.17\%.

In Figure \ref{fig_mnistServer}, we observe a similar loss in performance on the server models when using averaged weights. The model without pre-trained weights performs better than that with averaged weights. A similar observation is made while testing both models as shown in Table \ref{tab:mnist_test}. The model with averaged weights had a test accuracy of 72.76\%, precision of 77.97\%, recall of 72.76\%, and F1 score of 72.32\%. In comparison, the model without any pre-trained weights had a test accuracy of 79.38\%, precision of 81.28\%, recall of 79.38\%, and F1 score of 78.54\%. 

%% file: Contents/tables/mnist_testing.tex
\resizebox{\linewidth}{!}{
\begin{tabular}{llllll}
\hline \hline
\textbf{Device}   & \textbf{Setup}      & \textbf{Acc} & \textbf{Precision} & \textbf{Recall} & \textbf{F1 Score} \\
\hline
Edge 1          & Before update         & \textbf{0.9722}            & \textbf{0.9737}             & \textbf{0.9722}          & \textbf{0.9725}            \\
Edge 1          & After update            & 0.7544        & 0.6784       & 0.7544       & 0.6980              \\
\hline
Edge 2          & Before update         & \textbf{0.9568}            & \textbf{0.9594}             & \textbf{0.9568}          & \textbf{0.9571}            \\
Edge 2          & After update          & 0.5466         & 0.4836       & 0.5466        & 0.4617              \\
\hline
Server          & No pre-training           & \textbf{0.7938}   & \textbf{0.8128}    &\textbf{0.7938}  & \textbf{0.7854}      \\
Server          & Averaged weights          & 0.6417            & 0.6854             & 0.6417          & 0.6285   \\
\hline
\end{tabular}}

%% file: Contents/6_conclusion.tex
\section{Discussion}
\subsection{Performance improvement with ECAvg approach}
By fine-tuning on the server with a larger dataset, the model $\hat{M}$ solves a more challenging task with a higher number of classes, thereby improving generalisability of the model \cite{li2021cutpaste}. Updating the edge models with these parameters therefore results in better performance on their respective datasets. 

Models learn general features from the data during pre-training and their weights are initialized for fine-tuning. By averaging the weights of $M_1$ and $M_2$ on the server, the server's model $\hat{M}$ benefits from transfer learning to improve training performance. This improvement is evident in the fact that $\hat{M}$ with averaged weights has a better accuracy than an identical model pre-trained on a different dataset. The performance improvement is also consistent for precision, recall, and F1 score (see Experiment I \& II). 

The similarity between the local and global datasets provides common knowledge in the source and target tasks that is preserved in by averaging the weights. By exploiting this knowledge, the server model gains an advantage over an identical model whose weights are pre-trained on an unrelated task (ImageNet). This advantage is also retained downstream with the model update. The weights of the server model are fine-tuned on the global dataset and the knowledge is transferred to the edge models. The similarity of the tasks therefore results in the preservation of knowledge in the edge models that benefits the training process.


\subsection{Negative transfer learning with simple network}
The loss in performance is due to negative transfer learning \cite{rosenstein2005transfer}. Our method employs parameter transfer scheme \cite{lawrence2004learning}, which results in loss of shared knowledge when the source and target tasks are similar \cite{caoB2010adaptive}. When the source task is the edge datasets and the target task is the server dataset, loss in performance occurs on the target server model (as seen in Figure \ref{fig_mnistServer}. After the server updates the edge models with the fine tuned weights, the edge dataset becomes the target task and a similar loss in performance is observed (as seen in Figures \ref{fig_mnistModel1} and \ref{fig_mnistModel2}).

This negative transfer learning is avoided in the CIFAR-10 and CIFAR-100 tasks due to the more complex architecture of the MobileNetV2 and ResNet50 classifiers respectively. Training deep neural networks includes techniques such as regularization and hyperparameter tuning that optimize the weights for better performance. These techniques are not employed while training the neural network in the MNIST classification, resulting in negative transfer learning (see Experiment III).

\section{Conclusion and Future Works}
The use of edge and cloud together provides a collaboration between the two classes of devices in which the edge complements the shortcomings of the cloud and vice versa \cite{wang2020convergence}. Storage and memory constraints of edge devices can be overcome by offloading high storage demands and computationally intensive tasks to the server. 

In this paper, we proposed ECAvg - an edge-cloud collaborative learning approach in which edge devices perform pre-training of their local models on their respective datasets. The local datasets are and aggregated into a single large dataset on the server and the weights of the local models are also averaged to build a global model. The server then performs the computationally intensive task of training the global model on the much larger aggregated dataset. The local client models are then updated with the fine-tuned weights of the global model. Our approach differs from federated learning in that we avergae the weights and further fine-tune the global model on the server's dataset before updating the client models. 

We tested our approach with three experiments: MobileNetV2 on CIFAR-10 classification; ResNet50 on CIFAR-100 classification; and a simple convolutional neural network on MNIST classification. 

We observed significant performance improvement when using a deep learning model with our approach on the CIFAR-10 classification. Our server model with averaged weights had a better test accuracy (66.96\%) than a similar model with ImageNet weights (36.60\%). When the weights of the global model were transferred to the local edge models, their performance also improved on their respective datasets. The test accuracy of $M_1$ on Edge 1 improved from 38.86\% to 81.16\%, and the test accuracy of $M_2$ on Edge 2 improved from 36.88\% to 87.40\%. The performance improvements were also consistent across precision, recall, and F1 score for server and edge models. 

A similar observation was made with the CIFAR-100 classification, where the server model with averaged weights had a 37.22\% test accuracy, while a similar model with ImageNet weights had a 16.94\% test accuracy. Updating the edge model $M_1$ resulted in an increase in accuracy from 39.64\% to 51.56\%. Similarly, the accuracy of $M_2$ increased from 48.78\% to 51.80\% after the model update. 

A contrasting observation was made on the MNIST classification when a simple neural network architecture was used. The server model with averaged weights had a lower test accuracy (64.17\%) than a similar model without any pre-training (79.38\%). Similarly, the model updates resulted in a drop in performance on the edge models, where the test accuracy of $M_1$ dropped from 97.22\% to 75.44\% and the test accuracy of $M_2$ dropped from 95.68\% to 54.66\%. 

The models in MNIST suffer from negative transfer learning where the transfer of knowledge from the source task to the target task negatively affected the learning. This effect is not observed in the CIFAR-10 and CIFAR-100 classification tasks, where we used deep neural networks, and the models performed better using our approach. We conclude that deep neural networks mitigate negative transfer learning and thereby are essential to our proposed method. The performance improvements in the CIFAR-10 and CIFAR-100 classification shows the success of this method for deep learning.



We implemented our proposed method on two edge devices. However, this can be further extended to include any number of edge devices. We will also explore using different ratios for dividing the dataset into sub-datasets for the various number of edge devices.  


%% file: conference_101719.bbl
\begin{thebibliography}{10}

\bibitem{yao2022edge}
J.~Yao, S.~Zhang, Y.~Yao, F.~Wang, J.~Ma, J.~Zhang, Y.~Chu, L.~Ji, K.~Jia,
  T.~Shen, {\em et~al.}, ``Edge-cloud polarization and collaboration: A
  comprehensive survey for ai,'' {\em IEEE Transactions on Knowledge and Data
  Engineering}, vol.~35, no.~7, pp.~6866--6886, 2022.

\bibitem{sandler2018mobilenetv2}
M.~Sandler, A.~Howard, M.~Zhu, A.~Zhmoginov, and L.-C. Chen, ``Mobilenetv2:
  Inverted residuals and linear bottlenecks,'' in {\em Proceedings of the IEEE
  conference on computer vision and pattern recognition}, pp.~4510--4520, 2018.

\bibitem{tan2019efficientnet}
M.~Tan and Q.~Le, ``Efficientnet: Rethinking model scaling for convolutional
  neural networks,'' in {\em International conference on machine learning},
  pp.~6105--6114, PMLR, 2019.

\bibitem{bochkovskiy2020yolov4}
A.~Bochkovskiy, C.-Y. Wang, and H.-Y.~M. Liao, ``Yolov4: Optimal speed and
  accuracy of object detection,'' {\em arXiv preprint arXiv:2004.10934}, 2020.

\bibitem{liu2022research}
L.~Liu, C.~Ke, H.~Lin, H.~Xu, {\em et~al.}, ``Research on pedestrian detection
  algorithm based on mobilenet-yolo,'' {\em Computational intelligence and
  neuroscience}, vol.~2022, 2022.

\bibitem{chen2023edge}
C.~Chen, C.~Wang, B.~Liu, C.~He, L.~Cong, and S.~Wan, ``Edge intelligence
  empowered vehicle detection and image segmentation for autonomous vehicles,''
  {\em IEEE Transactions on Intelligent Transportation Systems}, 2023.

\bibitem{bibi2021edge}
R.~Bibi, Y.~Saeed, A.~Zeb, T.~M. Ghazal, T.~Rahman, R.~A. Said, S.~Abbas,
  M.~Ahmad, and M.~A. Khan, ``Edge ai-based automated detection and
  classification of road anomalies in vanet using deep learning,'' {\em
  Computational intelligence and neuroscience}, vol.~2021, pp.~1--16, 2021.

\bibitem{kukreja2019training}
N.~Kukreja, A.~Shilova, O.~Beaumont, J.~Huckelheim, N.~Ferrier, P.~Hovland, and
  G.~Gorman, ``Training on the edge: The why and the how,'' in {\em 2019 IEEE
  International Parallel and Distributed Processing Symposium Workshops
  (IPDPSW)}, pp.~899--903, IEEE, 2019.

\bibitem{cao2023fostering}
H.~Cao, M.~Wachowicz, R.~Richard, and C.-H. Hsu, ``Fostering new vertical and
  horizontal iot applications with intelligence everywhere,'' {\em arXiv
  preprint arXiv:2310.00346}, 2023.

\bibitem{yang2020big}
C.~Yang, S.~Lan, L.~Wang, W.~Shen, and G.~G. Huang, ``Big data driven
  edge-cloud collaboration architecture for cloud manufacturing: a software
  defined perspective,'' {\em IEEE access}, vol.~8, pp.~45938--45950, 2020.

\bibitem{hu2020coedge}
L.~Hu, G.~Sun, and Y.~Ren, ``Coedge: Exploiting the edge-cloud collaboration
  for faster deep learning,'' {\em IEEE Access}, vol.~8, pp.~100533--100541,
  2020.

\bibitem{wang2019adaptive}
S.~Wang, T.~Tuor, T.~Salonidis, K.~K. Leung, C.~Makaya, T.~He, and K.~Chan,
  ``Adaptive federated learning in resource constrained edge computing
  systems,'' {\em IEEE journal on selected areas in communications}, vol.~37,
  no.~6, pp.~1205--1221, 2019.

\bibitem{khan2020federated}
L.~U. Khan, S.~R. Pandey, N.~H. Tran, W.~Saad, Z.~Han, M.~N. Nguyen, and C.~S.
  Hong, ``Federated learning for edge networks: Resource optimization and
  incentive mechanism,'' {\em IEEE Communications Magazine}, vol.~58, no.~10,
  pp.~88--93, 2020.

\bibitem{su2021secure}
Z.~Su, Y.~Wang, T.~H. Luan, N.~Zhang, F.~Li, T.~Chen, and H.~Cao, ``Secure and
  efficient federated learning for smart grid with edge-cloud collaboration,''
  {\em IEEE Transactions on Industrial Informatics}, vol.~18, no.~2,
  pp.~1333--1344, 2021.

\bibitem{cao2019analytics}
H.~Cao and M.~Wachowicz, ``Analytics everywhere for streaming iot data,'' in
  {\em 2019 Sixth International Conference on Internet of Things: Systems,
  Management and Security (IOTSMS)}, pp.~18--25, IEEE, 2019.

\bibitem{cifar10}
A.~Krizhevsky, ``Learning multiple layers of features from tiny images,'' tech.
  rep., 2009.

\bibitem{lecun-mnisthandwrittendigit-2010}
Y.~LeCun and C.~Cortes, ``{MNIST} handwritten digit database,'' 2010.

\bibitem{li2021cutpaste}
C.-L. Li, K.~Sohn, J.~Yoon, and T.~Pfister, ``Cutpaste: Self-supervised
  learning for anomaly detection and localization,'' in {\em Proceedings of the
  IEEE/CVF conference on computer vision and pattern recognition},
  pp.~9664--9674, 2021.

\bibitem{rosenstein2005transfer}
M.~T. Rosenstein, Z.~Marx, L.~P. Kaelbling, and T.~G. Dietterich, ``To transfer
  or not to transfer,'' in {\em NIPS 2005 workshop on transfer learning},
  vol.~898, 2005.

\bibitem{lawrence2004learning}
N.~D. Lawrence and J.~C. Platt, ``Learning to learn with the informative vector
  machine,'' in {\em Proceedings of the twenty-first international conference
  on Machine learning}, p.~65, 2004.

\bibitem{caoB2010adaptive}
B.~Cao, S.~J. Pan, Y.~Zhang, D.-Y. Yeung, and Q.~Yang, ``Adaptive transfer
  learning,'' in {\em proceedings of the AAAI Conference on Artificial
  Intelligence}, vol.~24, pp.~407--412, 2010.

\bibitem{wang2020convergence}
X.~Wang, Y.~Han, V.~C. Leung, D.~Niyato, X.~Yan, and X.~Chen, ``Convergence of
  edge computing and deep learning: A comprehensive survey,'' {\em IEEE
  Communications Surveys \& Tutorials}, vol.~22, no.~2, pp.~869--904, 2020.

\end{thebibliography}
